\documentclass{article}
\usepackage{amsmath,graphicx,mlspconf}
\usepackage{xcolor}
\usepackage{amssymb}
\usepackage{mathtools}
\usepackage{booktabs}
\usepackage{multirow}
\usepackage{float}
\usepackage[inline]{enumitem}
\usepackage[capitalize,noabbrev]{cleveref}
\usepackage{placeins}
\usepackage{url}
\copyrightnotice{979-8-3195-0884-3/26/\$31.00 {\copyright}2026 IEEE}

\toappear{2026 IEEE International Workshop on Machine Learning for Signal Processing, Sep.\ 28-- Oct.\ 1, 2026, Atlanta, USA}

\title{ALAS: An Automatic Latent Alignment Score for Audio Language Models}
\name{Pooneh Mousavi$^{1,2}$, Yingzhi Wang$^3$, Mirco Ravanelli$^{1,2}$, Cem Subakan$^{4,1,2}$}
\address{$^1$Concordia University, $^2$Mila-Quebec AI Institute, $^3$Centrale Supelec, $^4$Laval University}

\begin{document}
\ninept
\maketitle

\begin{abstract}
\vspace{-.10cm}
Large Language Models (LLMs) are extended into Speech-LLMs, and the quality of the audio--text alignment they learn affects most downstream Spoken Language Understanding (SLU) behavior. Yet despite a growth of fusion strategies, there is no standard way to measure how well a Speech-LLM internally binds audio frames to text tokens. We introduce \textbf{ALAS} (Automatic Latent Alignment Score), a model- and task-agnostic metric that probes the LLM's per-layer hidden states, scoring the cross-modal cosine similarity between audio and text representations against a Whisper-derived reference. ALAS needs only a frozen forward pass and an off-the-shelf ASR reference, with no training or fitted classifier, and is calibrated to an interpretable uniform baseline comparable across tasks. Applying ALAS to four open-source Speech-LLMs (AF3, Qwen2-Audio, Qwen-Omni, SALMONN) across emotion recognition (IEMOCAP), open-ended SQA (LibriSQA), and multi-choice audio understanding (MMAU-speech), we find that the depth and strength of alignment reflect each model's audio-encoder design and the acoustic-versus-semantic demands of the task, and that ALAS tracks but does not duplicate task accuracy, exposing models that score well without genuinely grounding in the audio. We release ALAS as an open-source library so that practitioners can probe their own Speech-LLMs or try new tasks.
\vspace{-.1cm}
\end{abstract}


\newcommand{\cem}[1]{ {\color{blue} C: #1}}
\vspace{-.30cm}
\section{Introduction}
\label{sec:intro}
\vspace{-.10cm}

\begin{figure*}[!t]
  \centering
  \includegraphics[width=0.95\textwidth]{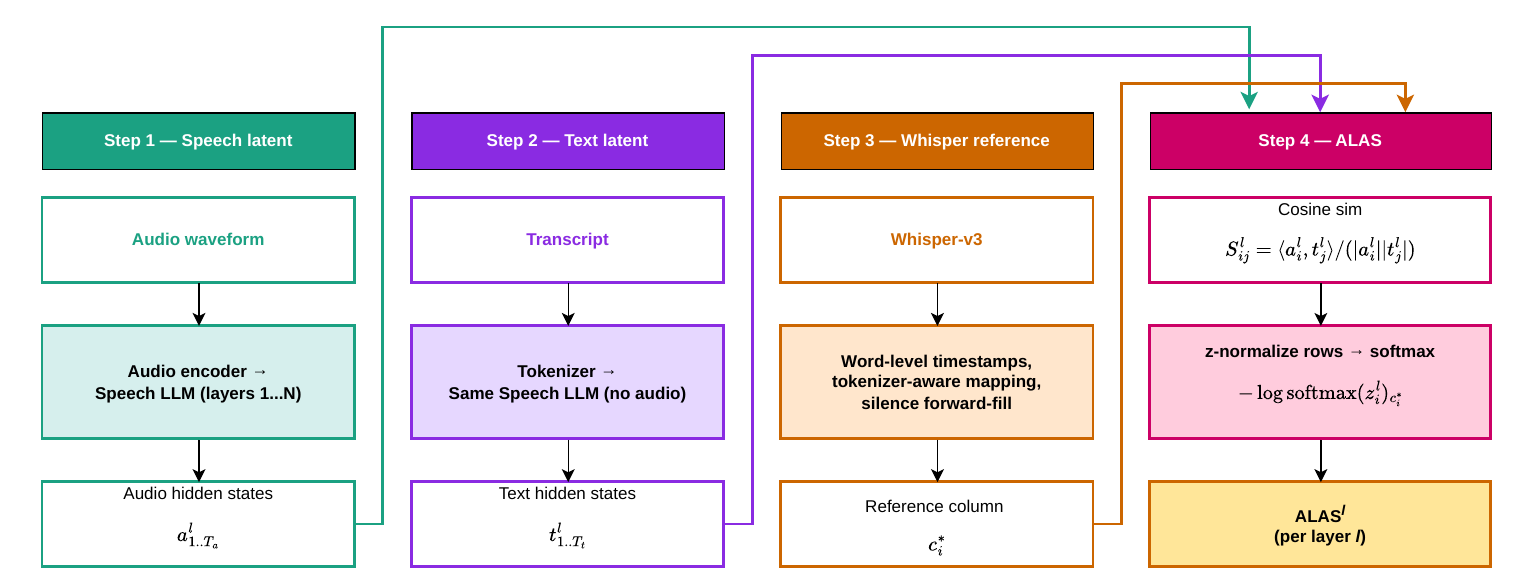}
  \vspace{-.20cm}
  \caption{The ALAS pipeline. \textbf{(1)} the Speech-LLM processes the audio waveform with a task prompt; we extract layer-wise audio hidden states. \textbf{(2)} the same LLM processes the transcript with the same prompt (no audio); we extract layer-wise text hidden states. \textbf{(3)} Whisper-v3 produces word-level timestamps which are mapped to LLM tokens, yielding a per-frame reference column. \textbf{(4)} the per-layer cosine-similarity row is $z$-normalized and the contrastive cross-entropy at the reference column produces $\mathrm{ALAS\text{-}soft}^{l}$.}
  \label{fig:alas_pipeline}
  \vspace{-.30cm}
\end{figure*}

Large Language Models (LLMs)~\cite{touvron2023llama, brown2020language} now power a growing family of multimodal models with image, speech, and audio inputs~\cite{tang2023salmonn, chu2024qwen2, li2023blip, mousavi2025listen}, with recent surveys synthesizing rapid progress on speech--language modeling~\cite{arora2025landscape, mousavi2025discrete,luo2026survey,mousavi2025investigating,gemini2023,rubenstein2023audiopalm,kumar2025competitive,zhang2026eureka}. A central design question for these models is how to \emph{fuse} the audio and text streams. How well the two are aligned governs whether the model reads the linguistic content of the speech, the words spoken, rather than falling back on its language prior, which is what most spoken-understanding tasks (spoken QA, instruction following, content-based reasoning) ultimately depend on. Alignment is not the whole story, since tasks like emotion recognition also need cues the transcript discards such as prosody; a good Speech-LLM must attend to both. But because text--audio alignment underpins the large class of transcript-dependent tasks, most current models invest heavily in it: the dominant recipe pairs a modality encoder with an adapter that projects encoder features into the LLM's embedding space~\cite{tan2024ssr}, while a complementary line pushes alignment into pre-training, via \emph{interleaved} speech--text decoding~\cite{maimon2025scaling} or self-generated descriptive pairs as in the DeSTA family~\cite{lu2025desta2}. Pre-training enforces alignment with contrastive, ASR, or interleaving objectives~\cite{chu2024qwen2,tang2023salmonn,maimon2025scaling}, and instruction tuning then \emph{assumes} it~\cite{lu2025desta2}. Yet how tightly a given model binds audio to text inside the LLM is almost never measured directly.

\noindent\textbf{Existing alignment tools fall short.} Traditional speech--text aligners such as MFA~\cite{mcauliffe2017montreal} and CTC-based methods~\cite{graves2006connectionist} operate on the audio--transcript pair externally; they cannot tell us whether the LLM has \emph{internally} learned the alignment or is bypassing it. Layer-wise probing of self-supervised speech encoders~\cite{pasad2021layer} characterizes phonetic and lexical content but not cross-modal binding, and cross-modal probing in vision--language~\cite{merullo2023linearly} does not address the speech--text setting where the audio side has a strict temporal order absent in image tokens. Closer to our setting, recent intervention-based probing tests whether an ALM uses provided acoustic cues in a grounded way~\cite{tsangko2026acoustic}, but it is confined to a single task (emotion recognition) and relies on hand-crafted, task-specific concept tokens, whereas ALAS is a task- and model-agnostic measurement of audio--text binding. The concurrent TARS framework~\cite{wang2026tars} is the most closely related effort: it explicitly couples \emph{representation alignment}, layer-wise hidden-state similarity between speech- and text-conditioned trajectories, with a behaviour-alignment reinforcement-learning signal to close the speech--text reasoning gap. Our work shares TARS's representation-alignment intuition but turns it into a model-agnostic \emph{diagnostic} metric rather than a training signal, and we use it to compare a heterogeneous set of open-source Speech-LLMs. Audio-LLM benchmarks~\cite{wang2024audiobenchuniversalbenchmarkaudio,zhao2023librisqa,sakshi2024mmau} report task accuracy, which collapses two distinct failure modes into one number: a model can score well by exploiting language priors while ignoring the audio, or by relying on acoustic shortcuts that never bind to text. What is missing is a \emph{model-agnostic, post-hoc} measurement of how tightly audio frames bind to transcript tokens inside the LLM, and how design choices (linear vs heavier adapter, Q-Former, interleaved pre-training) reshape that binding through the transformer stack.

\noindent\textbf{Contributions.} We close this gap with the following:
\begin{itemize}[noitemsep,topsep=0pt,leftmargin=*]
\item We propose \textbf{ALAS}, a probing metric that scores a Speech-LLM's per-layer cross-modal alignment using only the model's frozen forward pass and an off-the-shelf ASR reference. ALAS is model- and task-agnostic, has an interpretable uniform reference that makes it cross-task comparable, and requires no manual alignment or fitted classifier.
\item Across four open-source Speech-LLMs (Audio Flamingo 3 (AF3)~\cite{kong2025audioflamingo3}, Qwen2-Audio~\cite{chu2024qwen2}, Qwen-Omni~\cite{xu2025qwen25omni}, SALMONN~\cite{tang2023salmonn}) and three benchmarks (IEMOCAP~\cite{busso2008iemocap}, LibriSQA~\cite{zhao2023librisqa}, MMAU-speech~\cite{sakshi2024mmau}), we show that the depth at which alignment peaks is consistent with each model's audio-adapter design and encoder-training recipe, that a mismatched-transcript control cleanly separates models with genuine transcript-specific binding from one without it, and that, read alongside task accuracy, ALAS distinguishes models that ``listen'' from models that ``guess''.
\item We release ALAS as an easy-to-use open-source library: scoring a new Speech-LLM takes only a frozen forward pass and a transcript, with no retraining or per-model tuning, so practitioners can probe the alignment of their own models out of the box.\footnote{Code: \url{https://github.com/poonehmousavi/alas}}
\end{itemize}

\vspace{-.25cm}
\section{Methodology}
\label{sec:method}
\vspace{-.10cm}

The ALAS pipeline (Fig.~\ref{fig:alas_pipeline}) consists of four steps: extracting speech and text latents from the Speech-LLM, obtaining a reference alignment from Whisper, and computing the alignment score from these three inputs. The pipeline is fully automatic, model-agnostic, and requires no training or manual alignment.

\noindent\textbf{Speech and text latents (Steps 1--2).} For each utterance we run the Speech-LLM twice: once on (audio + task instruction) and once on (transcript + same instruction, with no audio). The text branch is fed the \emph{Whisper transcript} of the same audio, the same source used for the reference alignment in Step~3; using one consistent ASR transcript on both the text branch and the reference keeps them in register, so any ASR errors affect both identically and do not introduce a spurious audio--text mismatch into the score. From each pass we extract the per-layer hidden states and trim them to the modality-specific span (audio frames on the audio pass, transcript tokens on the text pass). The result is a tensor of shape $(L, T_{\text{audio}}, d)$ for the audio branch and $(L, T_{\text{text}}, d)$ for the text branch, where $L$ is the number of transformer layers and $d$ the hidden dimension. 

% Unless stated otherwise we report ALAS on the full task split (the \emph{all} column of Table~\ref{tab:alas_results}); the gold-correct audio$\checkmark$\,text$\checkmark$ subset and the response-agreement tertiles (Sec.~\ref{sec:exp}) are auxiliary partitions of the same samples, not a pre-filter on the metric.

\noindent\textbf{Reference alignment from Whisper (Step 3).} Whisper-v3~\cite{radford2022robustspeechrecognitionlargescale} provides, for each spoken word $w$, a time window $[s_w, e_w)$ during which it is uttered. Each audio frame $i$ is assigned to the word whose window contains its timestamp $\tau_i$, i.e.\ the word $w$ with $s_w \le \tau_i < e_w$. We then identify the corresponding target on the text side: the index $j$ of the LLM token that frame $i$ should align with. Since the tokenizer frequently splits a word into several sub-tokens, we set the reference column to the \emph{first} sub-token of that word, $c^*_i = j$, and count a frame as correct (for ALAS-hard) if any sub-token of the word is ranked highest, so that the score is not penalized merely for a word being split. Frames falling in a silence between two words (no window contains $\tau_i$) inherit the most recently uttered word, so that pausing on a word is likewise not penalized. This yields, for every audio frame $i$, the index $c^*_i$ of its target text token, which is used in the score below.

\noindent\textbf{Alignment score (Step 4).} Let $a^l_i \in \mathbb{R}^d$ denote the LLM's hidden state at audio frame $i \in \{1,\ldots,T_{\text{audio}}\}$ and $t^l_j \in \mathbb{R}^d$ the hidden state at text token $j \in \{1,\ldots,T_{\text{text}}\}$, both at transformer layer $l$. We define the per-layer cosine-similarity matrix $S^l \in \mathbb{R}^{T_{\text{audio}} \times T_{\text{text}}}$ and its row-wise $z$-normalization $z^l$ (with row mean $\mu^l_i$ and standard deviation $\sigma^l_i$) as
\vspace{-.10cm}
\begin{equation}
\label{eq:sim}
S^l_{i,j} = \frac{\langle a^l_i, t^l_j\rangle}{\|a^l_i\|\,\|t^l_j\|},
\qquad
z^l_{i,j} = \frac{S^l_{i,j} - \mu^l_i}{\sigma^l_i + \epsilon}.
\end{equation}
The $z$-normalization removes the layer- and model-specific scale of the cosine matrix, so the metric depends only on the \emph{relative} ranking of text tokens for each audio frame, not on the absolute magnitude of the embeddings, which is known to vary widely across layers and models~\cite{ethayarajh2019contextual}. ALAS therefore asks a single scale-free question per audio frame: does its hidden state rank the reference text token above the others? This makes scores comparable across layers, models, and (after the $\log T_{\text{text}}$ scaling introduced below) tasks.
 
From this per-frame ranking we compute two scores, both averaged over the \emph{active set} $\mathcal{A}^*$ of audio frames inside a Whisper word window (silence-only frames are excluded so long silences cannot dominate). \textbf{ALAS-soft} is the per-row contrastive cross-entropy at the reference column $c^*_i$ {(as defined in Step 3.)},
\vspace{-.10cm}
\begin{equation}
\label{eq:alas}
\mathrm{ALAS\text{-}soft}^l = \frac{1}{|\mathcal{A}^*|}\!\!\sum_{i \in \mathcal{A}^*}\!\! -\log \frac{\exp\!\left(z^l_{i,c^*_i}\right)}{\sum_{j=1}^{T_{\text{text}}} \exp\!\left(z^l_{i,j}\right)}.
\end{equation}
For audio frame $i$, the softmax turns its similarity row $z^l_{i,\cdot}$ into a distribution over the $T_{\text{text}}$ text tokens, and ALAS-soft is the negative log-probability it assigns to the correct token $c^*_i$, averaged over frames: small when most of the mass lands on the reference token, large when it spreads out or peaks on the wrong one. A row equally similar to all tokens gives each probability $1/T_{\text{text}}$ and contributes $\log T_{\text{text}}$, so dividing by $\log T_{\text{text}}$ puts this no-information case at exactly $1$. The scaled score reads off against that baseline: below $1$ the hidden states point at the reference token (better than chance), near $1$ there is little token-level signal, and above $1$ they favour the wrong token. Each utterance is normalized by its own $T_{\text{text}}$, so the baseline is exact and needs no calibration.
 
\textbf{ALAS-hard} replaces the softmax with the rank of the reference token,
\vspace{-.10cm}
\begin{equation}
\label{eq:alashard}
\mathrm{ALAS\text{-}hard}^l = \frac{1}{|\mathcal{A}^*|}\sum_{i \in \mathcal{A}^*} \frac{1}{\mathrm{rank}^l_i},
\end{equation}
where $\mathrm{rank}^l_i$ is the position of $c^*_i$ once tokens are sorted by $z^l_{i,\cdot}$ (the reference column's mean reciprocal rank). The two are complementary: ALAS-soft stays low when the correct token is merely among the top few, whereas ALAS-hard credits only frames that pin it exactly. A large soft-vs-hard gap thus flags a model that aligns to the right \emph{region} of the transcript without resolving the exact token, the signature of SALMONN's window-pooled Q-Former (Sec.~\ref{sec:exp}).

% \noindent\textbf{Embedding-layer Whisper-self-consistency.} The four models we benchmark all use Whisper-derived audio encoders, and we feed the text branch the Whisper transcript. At the embedding layer, both branches therefore inherit Whisper's own representation space, and the cosine similarity is trivially high at corresponding positions regardless of LLM behaviour. This produces an anomalously low $\mathrm{ALAS\text{-}soft}^0$ that is not a property of the LLM. We report all results from layer $L \ge 3$; the artifact magnitude is comparable to inter-model differences and large enough to swap the ranking of two adjacent models if not corrected.

\vspace{-.20cm}
\section{Experiments}
\label{sec:exp}
\vspace{-.10cm}

\noindent\textbf{Models.} We benchmark four open-source Speech-LLMs whose audio-encoder, adapter, and training recipes span the following design spectrum. Qwen2-Audio-7B-Instruct~\cite{chu2024qwen2} is the lightest of the four: it feeds Whisper-large-v3 features (stride-2 pooled to ${\approx}25$\,Hz) directly into Qwen-7B with \emph{no} learned adapter, and trains the Whisper encoder jointly with the LLM only during pre-training (frozen at instruction-tuning). Qwen2.5-Omni-7B Thinker~\cite{xu2025qwen25omni} adds a learned projection adapter between a streaming-modified Whisper encoder and a Qwen2.5-based Thinker LLM; encoder and LLM are co-tuned in two stages. AF3~\cite{kong2025audioflamingo3} replaces vanilla Whisper with \emph{AF-Whisper} (a Whisper-v3 backbone further trained on speech, sound, and music captioning) and connects it to Qwen2.5-7B through projector adapter layers; the encoder is tuned in stage 2 then frozen. SALMONN-7B~\cite{tang2023salmonn} sits at the opposite end: two \emph{frozen} encoders (Whisper-v2 + BEATs) are bridged to Vicuna-13B by a window-level Q-Former (one query per $0.33$\,s window, ${\approx}3$\,Hz output rate) plus LoRA on the LLM. Together these four span the design axis from ``no adapter, encoder co-tuned with LLM'' to ``heavy compression-based adapter, encoders frozen'', which is precisely the axis our layer-depth analysis (Sec.~\ref{sec:result1}) tracks

\noindent\textbf{Tasks.} We choose three benchmarks that stress different aspects of speech understanding: IEMOCAP~\cite{busso2008iemocap}, four-way emotion recognition (acoustic / paralinguistic); LibriSQA~\cite{zhao2023librisqa}, free-form spoken question answering (semantic / linguistic); and the speech subset of MMAU~\cite{sakshi2024mmau}, four-way multi-choice audio understanding (reasoning + grounded selection). Together they probe how alignment generalises across acoustic, semantic, and reasoning workloads.

\noindent\textbf{Evaluation.} Correctness is scored against gold by keyword emotion matching (IEMOCAP), sentence-transformer cosine $\ge 0.7$ (LibriSQA), and exact letter match (MMAU). We run each model twice per sample, an \emph{audio branch} (audio + instruction) and a \emph{text branch} (Whisper transcript + instruction, no audio), and grade each, giving per-branch accuracies \textbf{a-acc} and \textbf{t-acc}. We report ALAS both on the full split (\emph{all}) and on the subset where both branches answer correctly ($n_{\checkmark\checkmark}$).

\begin{table}[!t]
\centering
\scriptsize
\setlength{\tabcolsep}{2.0pt}
\renewcommand{\arraystretch}{0.96}
\begin{tabular}{@{}llcccccccc@{}}
\toprule
\multirow{2}{*}{Task} & \multirow{2}{*}{Model} & \multirow{2}{*}{$n_{\checkmark\checkmark}$} & \multirow{2}{*}{a-acc} & \multirow{2}{*}{t-acc} & \multirow{2}{*}{$L^*$} & \multicolumn{2}{c}{ALAS-soft$\downarrow$} & \multicolumn{2}{c}{ALAS-hard$\uparrow$} \\
\cmidrule(lr){7-8}\cmidrule(lr){9-10}
   &  &   &  &  &  & $\checkmark\checkmark$ & all & $\checkmark\checkmark$ & all \\
\midrule
\multirow{4}{*}{IEMOCAP}
        & AF3                & 222 & 0.73 & 0.44 & 28 & 1.03 & 1.04 & 0.383 & 0.378 \\
        & Qwen2-Audio        & 190 & 0.59 & 0.46 & 16 & 1.05 & 1.05 & 0.396 & 0.398 \\
        & \textbf{Qwen-Omni} & \textbf{231} & 0.64 & 0.51 & 27 & \textbf{1.00} & \textbf{0.99} & \textbf{0.433} & \textbf{0.436} \\
        & SALMONN            & 183 & 0.50 & 0.35 & 18 & 1.04 & 1.10 & 0.398 & 0.367 \\
\midrule
\multirow{4}{*}{LibriSQA}
         & \textbf{AF3}      & \textbf{1750} & \textbf{0.80} & \textbf{0.72} & 20 & \textbf{0.95} & \textbf{0.97} & \textbf{0.385} & \textbf{0.376} \\
         & Qwen2-Audio       & 1174 & 0.57 & 0.56 & 32 & 1.00 & 1.02 & 0.369 & 0.362 \\
         & Qwen-Omni         & 1167 & 0.51 & 0.56 & 22 & 1.02 & 1.04 & 0.352 & 0.340 \\
         & SALMONN           & 1382 & 0.67 & 0.64 & 38 & 1.14 & 1.14 & 0.202 & 0.204 \\
\midrule
\multirow{4}{*}{MMAU-speech}
         & AF3                & 176 & 0.67 & 0.66 & 27 & 0.95 & 0.92 & 0.293 & 0.316 \\
         & \textbf{Qwen2-Audio} & 128 & 0.54 & 0.55 & 32 & \textbf{0.91} & \textbf{0.90} & 0.307 & 0.319 \\
         & \textbf{Qwen-Omni} & \textbf{201} & \textbf{0.76} & \textbf{0.76} & 22 & \textbf{0.91} & \textbf{0.90} & \textbf{0.311} & \textbf{0.321} \\
         & SALMONN            & 102 & 0.41 & 0.53 & 35 & 1.10 & 1.10 & 0.148 & 0.147 \\
\bottomrule
\end{tabular}
\vspace{-.05cm}
\caption{ALAS scores for each model, reported at its best layer $L^*$ (the layer with the lowest ALAS-soft, searched from layer $3$ onward to skip the embedding-layer artifact). \textbf{ALAS-soft} is the scaled cross-entropy of Eq.~\ref{eq:alas} (lower is better; $1.0$ is the uninformative baseline); \textbf{ALAS-hard} is the rank-1 score (higher is better). Both are given on the subset where both branches answer correctly ($\checkmark\checkmark$, $n_{\checkmark\checkmark}$ samples) and on the full split (\emph{all}, $n\!=\!529, 2618, 281$); the \emph{all} column is comparable across models. a-acc and t-acc are the audio- and text-branch accuracies on the full split. The best model per task is in bold; on MMAU the two bolded ALAS-soft leaders have overlapping confidence intervals.}
\label{tab:alas_results}
\vspace{-.40cm}
\end{table}

\begin{figure*}[!t]
  \centering
  \begin{minipage}[t]{0.32\textwidth}\centering
    \includegraphics[width=\textwidth]{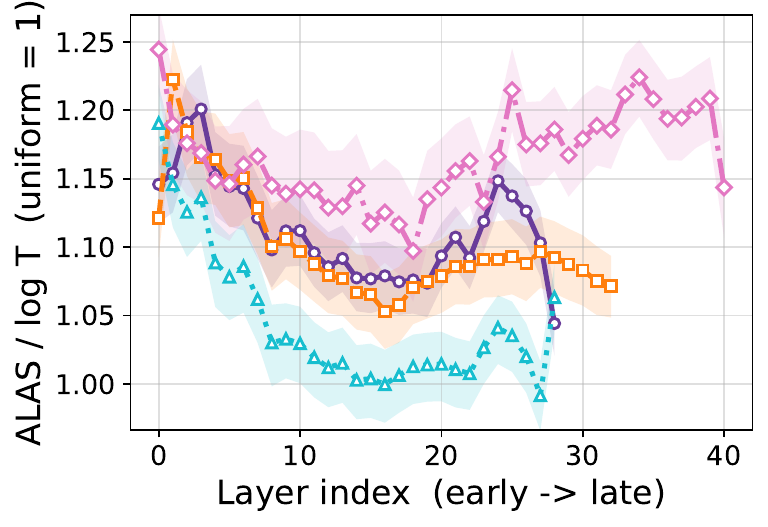}\\[-0.30em]\small{(a) IEMOCAP}
  \end{minipage}\hfill
  \begin{minipage}[t]{0.32\textwidth}\centering
    \includegraphics[width=\textwidth]{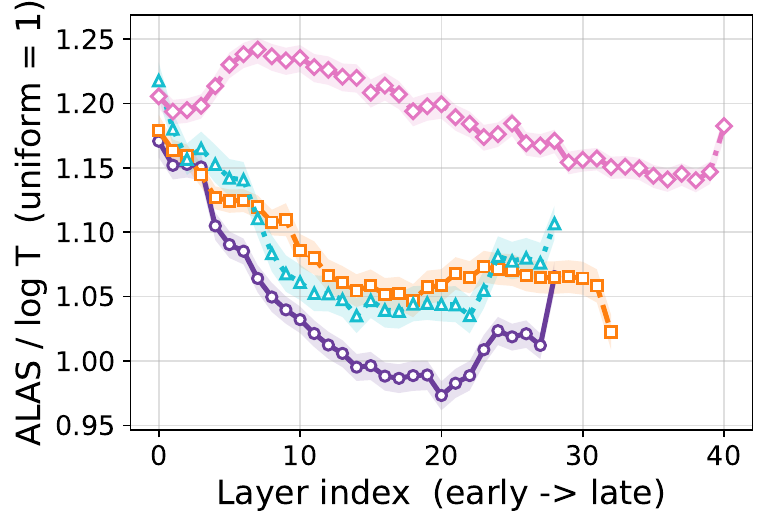}\\[-0.30em]\small{(b) LibriSQA}
  \end{minipage}\hfill
  \begin{minipage}[t]{0.32\textwidth}\centering
    \includegraphics[width=\textwidth]{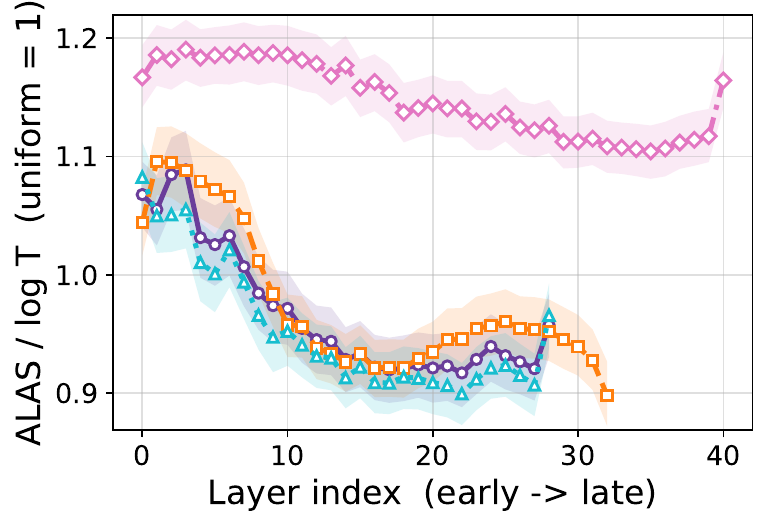}\\[-0.30em]\small{(c) MMAU-speech}
  \end{minipage}\\[0.30em]
  \includegraphics[width=0.55\textwidth]{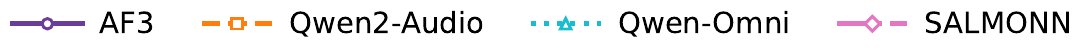}
  \vspace{-.15cm}
  \caption{ALAS-soft ($/\!\log T_{\text{text}}$, lower is better) per LLM transformer layer for the four models on the three benchmarks. Shaded bands are 95\% bootstrap CIs around the per-layer mean ($n=$ full split). The uniform-row baseline sits at $1.0$. The location of each model's minimum is consistent with its audio-adapter design (Sec.~\ref{sec:exp}).}
  \label{fig:quantitative}
  \vspace{-.30cm}
\end{figure*}

\section{Results}
\label{sec:result}
\subsection{ALAS complements task accuracy and tracks the training recipe}
\label{sec:result1}

\vspace{-.05cm}

ALAS and downstream accuracy capture related but distinct aspects of a model, and are therefore complementary rather than redundant. The best-aligned model is typically near, though not necessarily at, the top on audio accuracy: the lowest-ALAS-soft model on each benchmark is Qwen-Omni on IEMOCAP ($0.99$, a-acc $0.64$), AF3 on LibriSQA ($0.97$, a-acc $0.80$), and the Qwen pair on MMAU ($0.90$). We do not interpret this as evidence that alignment determines accuracy; rather, we read the per-task winners as consistent with the audio-encoder--LLM training recipe. From these four models we hypothesise that Qwen-Omni's jointly co-tuned encoder suits the prosodic cues of emotion, while AF3's caption-trained AF-Whisper encoder provides the richer semantic features that benefit free-form QA.
 
Where the two diverge is the more informative signal. On MMAU, Qwen-Omni and Qwen2-Audio reach identical ALAS-soft ($0.90$) yet differ sharply in accuracy ($0.76$ vs $0.54$): equally tight internal alignment, very different task success, which a single accuracy number would conflate. The gap between the two ALAS variants is similarly diagnostic. ALAS-soft credits a frame for placing the reference token among its top matches, whereas ALAS-hard requires it to rank first; a large soft--hard gap thus marks alignment to the right transcript \emph{region} without exact-token resolution. SALMONN is the clear case: on the semantic tasks its ALAS-soft is only moderately poor ($1.14$, $1.10$) but its ALAS-hard collapses ($0.20$, $0.15$, versus $0.34$--$0.44$ for the others), exactly the footprint of its $\approx 3$\,Hz Q-Former, which preserves coarse ordering but destroys token-level precision. On IEMOCAP, where the target is a single emotion word and no fine token alignment is needed, SALMONN's ALAS-hard ($0.40$) is unremarkable, confirming the gap is task-dependent rather than a constant model defect. Finally, the count of jointly-correct samples $n_{\checkmark\checkmark}$ tracks ALAS-soft within each task (e.g.\ AF3 leads LibriSQA with $n_{\checkmark\checkmark}=1750$, SALMONN trails MMAU with $102$), so tighter alignment co-occurs with more samples both branches solve. The alignment--accuracy association is directional but underpowered at this scale (pooled Spearman $\rho\!=\!-0.43$, $n\!=\!12$, not significant), so we treat it as suggestive and rest the validity of ALAS on the mismatched-transcript control below.
 
\vspace{-.05cm}
\subsection{Alignment depth varies with the adapter and encoder-training recipe}
\vspace{-.05cm}

Having established that ALAS captures meaningful per-model differences, we now ask \emph{where} in the network each model achieves its alignment, read from the best-layer column $L^*$ and the per-layer curves of Fig.~\ref{fig:quantitative}. The depth of the alignment peak orders the models along a single architectural axis: how much work the LLM stack must do to bind audio to text, which in turn depends on the adapter and on whether the encoder is trained with the LLM. Qwen-Omni, whose projection adapter is co-tuned with the Thinker LLM, presents features the model can use almost immediately and peaks earliest ($L^*\!=\!22$ on both semantic tasks). Qwen2-Audio feeds pooled Whisper features in with no learned adapter, so the stack itself must align them and the peak moves later ($L^*\!=\!32$ on the semantic tasks). AF3's heavy, caption-trained adapter delivers rich but aggressively reshaped features, requiring most of the network's depth to re-bind ($L^*\!=\!20$--$28$). SALMONN is the extreme: with both encoders frozen and a $\approx 3$\,Hz Q-Former discarding most temporal structure, the LLM must rebuild alignment almost from scratch through LoRA-adapted attention, and peaks deepest of all ($L^*\!=\!38$, $35$). The one departure is SALMONN's mid-stack peak on IEMOCAP ($L^*\!=\!18$): emotion is recoverable from acoustic cues already present in its BEATs branch, so it needs no late-stage token re-binding, the same task structure visible in the absolute-level analysis below.

These curves are broad plateaus rather than sharp optima, so $L^*$ should be read as the centre of a stable low-ALAS region, not a single privileged layer. Because the four models also differ in base LLM and depth, we frame this depth ordering as consistent with audio-encoder design rather than a controlled attribution; the mismatched-transcript control isolates the alignment signal from these confounds.

The absolute \emph{level} of the best ALAS-soft adds an orthogonal, task-driven reading. It splits cleanly by whether a task is solved from the words or from the sound: on IEMOCAP, where emotion lives in prosody, the best score stays at or above the uniform baseline of $1.0$ for three of four models (weak token-level binding, as expected), whereas on MMAU, which demands grounding in the transcribed content, three of four reach clearly sub-uniform binding ($\le 0.92$). LibriSQA sits between the two. The same metric therefore exposes \emph{where} a model aligns (depth, set by architecture) and \emph{how strongly} it must (level, set by the task).
 
\vspace{-.05cm}
\subsection{Qualitative structure of the learned alignments}
\vspace{-.05cm}
 
\begin{figure*}[!t]
  \centering
  \includegraphics[width=\textwidth]{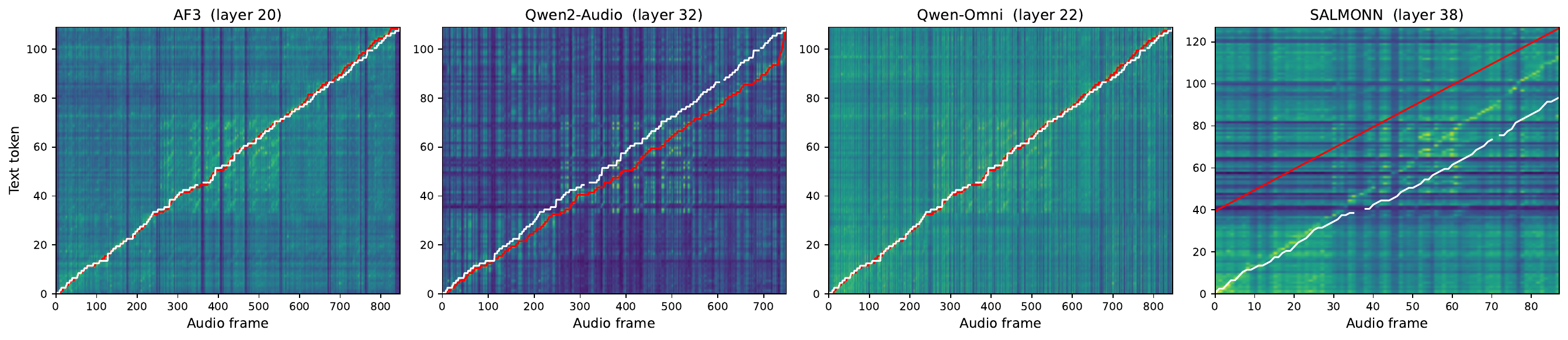}
  \vspace{-.25cm}
  \caption{Cross-modal similarity heatmaps for the same long-audio LibriSQA utterance (${\sim}17$\,s, 109 LLM text tokens for AF3) across all four models at each model's best layer. \emph{White}: Whisper reference alignment; \emph{red}: the MAS path recovered from the model's own similarity matrix.}
  \label{fig:heatmaps}
  \vspace{-.30cm}
\end{figure*}
 
Fig.~\ref{fig:heatmaps} shows, for one LibriSQA utterance, the cross-modal similarity between audio frames and text tokens at each model's best layer. On top of each heatmap we draw the model's own alignment path, recovered by monotonic alignment search (MAS)~\cite{kim2020glowttsgenerativeflowtexttospeech}: using dynamic programming, MAS finds the path through the similarity matrix that maximizes the total similarity along it while moving monotonically (each successive audio frame maps to the same or a later text token), giving the best non-crossing frame-to-token assignment. We compare this path against the Whisper reference. The four models bind audio to text with visibly different sharpness. AF3 produces a tight, continuous diagonal whose MAS path closely follows the reference. Qwen2-Audio's diagonal is noisier, with off-diagonal energy as its later layers turn more semantic. Qwen-Omni is sharp early but diffuses in the second half, consistent with its design abstracting away fine temporal detail. SALMONN's audio axis is far shorter than the others ($\sim$88 vs $\sim$800 frames), a direct visual consequence of its Q-Former compression; its path still follows the reference but only at a coarse granularity. These views match the quantitative picture: binding sharpens as the transformer stack mixes the two modalities.

\vspace{-.05cm}
\subsection{Mismatched-transcript control}
\vspace{-.05cm}
 
As a control that ALAS captures genuine binding rather than mean hidden-state structure, we re-score each utterance's audio by pairing its frame embeddings with the text-branch embeddings of a \emph{different} randomly-chosen utterance. If ALAS measures real binding, the mismatched score should rise toward the uniform baseline; if not, it should match the original. Table~\ref{tab:mismatch} reports the resulting matched-versus-mismatched gap per (model, task).
 
\begin{table}[!h]
\centering
\footnotesize
\setlength{\tabcolsep}{6pt}
\renewcommand{\arraystretch}{0.98}
\begin{tabular}{@{}lccc@{}}
\toprule
Model & IEMOCAP & LibriSQA & MMAU \\
\midrule
AF3         & $+0.08$          & $\mathbf{+0.23}$ & $\mathbf{+0.10}$ \\
Qwen2-Audio & $\mathbf{+0.10}$ & $\mathbf{+0.20}$ & $\mathbf{+0.12}$ \\
Qwen-Omni   & $+0.08$          & $\mathbf{+0.18}$ & $\mathbf{+0.12}$ \\
SALMONN     & $-0.02$          & $-0.03$          & $+0.01$ \\
\bottomrule
\end{tabular}
\vspace{-.05cm}
\caption{Mismatched-transcript control: the change in ALAS-soft at each model's best layer when the text branch is paired with a random other utterance's transcript ($n\!=\!80$ pairs/cell). A positive gap means the matched transcript scores better than a random one; i.e., ALAS responds to the specific text identity. \textbf{Bold} marks gaps whose 95\% bootstrap CI excludes $0$ (CI half-widths are $0.07$--$0.11$). SALMONN's gap is statistically indistinguishable from zero on all three tasks, directly confirming that its hidden states carry no transcript-specific signal.}
\label{tab:mismatch}
\vspace{-.30cm}
\end{table}
 
The result is clear (Table~\ref{tab:mismatch}): for AF3, Qwen2-Audio, and Qwen-Omni the matched transcript scores meaningfully better than a random one, so ALAS responds to the specific transcript and not to generic hidden-state structure. SALMONN is the exception, with no significant gap on any task, which independently confirms that its frozen-encoder, heavily-compressed pipeline retains no transcript-specific signal. This control is our primary evidence that ALAS measures real audio--text binding.
 
\vspace{-.05cm}
\subsection{Internal response agreement predicts alignment}
\vspace{-.05cm}
 
\begin{figure}[!t]
  \centering
  \includegraphics[width=0.98\columnwidth]{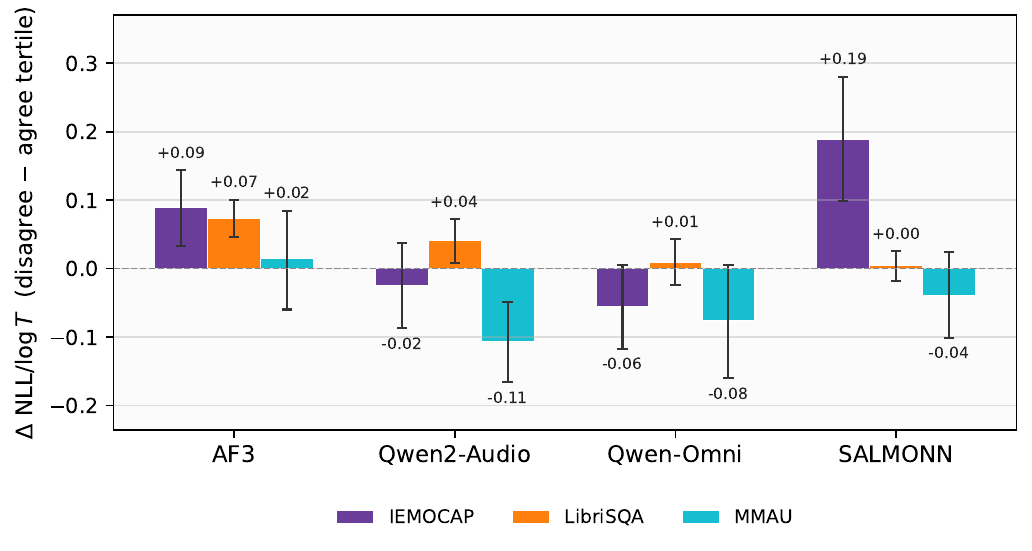}
  \vspace{-.20cm}
  \caption{Disagree-tertile minus agree-tertile ALAS-soft at each model's best layer, 95\% bootstrap CI as error bars. Positive (above the dashed line) means agreement tracks alignment (the LibriSQA pattern for the trainable-encoder models and the SALMONN--IEMOCAP signature); negative means disagreement tracks alignment (the MMAU pattern); a CI crossing zero means no resolvable signal.}
  \label{fig:buckets}
  \vspace{-.30cm}
\end{figure}

As a final, gold-free check we ask whether ALAS tracks a model's own behaviour: when its audio and text branches give the same answer, is its internal alignment tighter than when they disagree? For each (task, model) we split samples into three equal groups by how similar the two branch answers are (\emph{agree}, \emph{neutral}, \emph{disagree}, by SBERT~\cite{reimers-2019-sentence-bert} cosine) and compare ALAS-soft across them. Fig.~\ref{fig:buckets} summarises each cell as the disagree-minus-agree difference.
 
The direction of this effect depends on the task. On LibriSQA, where the answer is the spoken content, alignment is tighter when the two branches agree: for AF3 and Qwen2-Audio ALAS-soft rises monotonically from the agree to the disagree group, with non-overlapping confidence intervals. On IEMOCAP the effect is weak, as expected for a paralinguistic task that needs little token-level binding. On MMAU it reverses: alignment is tightest in the \emph{disagree} group, because the questions that truly require listening, where the audio answer departs from the transcript-only answer, are exactly the ones where the audio branch must bind closely to the speech. In every case SALMONN shows no resolvable effect, consistent with its hidden states carrying no transcript-specific signal. The same metric thus serves as a behavioural diagnostic, with the sign of the agreement effect reflecting whether a task is solved from the words, from prosody, or from genuine listening.
 
\vspace{-.20cm}
\section{Conclusions}
\label{sec:conclusion}
\vspace{-.10cm}
 
Audio--text alignment underpins Speech-LLM behavior yet is rarely measured directly. We introduced ALAS, a cross-modal alignment metric computed from the LLM's frozen forward pass with no training or fitted classifier. Across four open-source Speech-LLMs and three benchmarks, ALAS recovers interpretable layer-wise patterns whose depth and level are consistent with each model's adapter and encoder-training recipe, and a mismatched-transcript control confirms that it responds to the specific transcript rather than to incidental hidden-state geometry. Read alongside task accuracy, ALAS separates models that ground their answers in the audio from those that lean on language priors, making it a useful complement on speech-LLM leaderboards. We release ALAS as a lightweight, pip-installable library so practitioners can apply it to their own models out of the box. Multilingual evaluation and using ALAS as an auxiliary alignment objective during pretraining are natural next steps.
 
\FloatBarrier

 \section{Acknowledgment}
 We acknowledge the support of the Natural Sciences and Engineering Research Council of Canada (NSERC) and the Digital Research Alliance of Canada (alliancecan.ca). Cem Subakan is supported by the NSERC project RGPIN-2023-05759.
\bibliographystyle{IEEEbib}
\bibliography{strings,refs}
 
\end{document}